\documentclass{article} 
\usepackage{iclr2025_conference,times}

\usepackage{amsmath,amsfonts,bm}

\def\eqref#1{equation~\ref{#1}}

\def\1{\bm{1}}

\DeclareMathAlphabet{\mathsfit}{\encodingdefault}{\sfdefault}{m}{sl}
\SetMathAlphabet{\mathsfit}{bold}{\encodingdefault}{\sfdefault}{bx}{n}

\usepackage{hyperref}
\usepackage{url}
\usepackage{authblk}
\usepackage{graphicx}

\title{Diagnosing Low-Confidence Patient Profiles via Localized Harmful Covariate Shift Detection} 

\author[1]{Olivier Lefebvre}
\author[1]{Hakima Laribi}
\author[2,3]{Dan Poenaru}
\author[4]{Martin Vallières}

\affil[1]{Department of Computer Science, Université de Sherbrooke, Sherbrooke, Canada}
\affil[2]{Department of Pediatric Surgery, Montreal Children’s Hospital, Montreal, Canada}
\affil[3]{Centre for Outcomes Research and Evaluation (CORE), Research Institute of the McGill University Health Centre, Montreal, Canada}
\affil[4]{Medical Physics Unit, Department of Oncology, McGill University, Montreal, Canada}

\iclrfinalcopy 
\begin{document}

\maketitle

\begin{abstract}

Before deploying clinical prediction models, their reliability must be assessed to ensure safe use. Errors may arise from distribution shifts between training and deployment or from heterogeneous performance where the model systematically underperforms for certain patient subgroups. Yet most post‑hoc diagnostic methods evaluate reliability only at the population level, checking whether the deployment data resemble the training data and whether the model performs well on average, without accounting for heterogeneity that may cause some patients to be disproportionately affected.
We propose a unified diagnostic pipeline that combines both perspectives. MED3pa first discovers interpretable low‑confidence patient profiles, and Detectron is then applied within each profile to test whether its degradation is associated with a detectable harmful covariate shift.
Across three clinical prediction tasks and three injected subgroup shifts in MIMIC‑IV, we find that shift detectability itself is informative. An injected age shift, undetectable at the population level, was nonetheless localized to specific degraded profiles, whereas sex‑ and race‑related shifts were detected broadly and therefore offered little guidance for identifying the most affected patients.

\end{abstract}

\section{INTRODUCTION}

Machine learning (ML) models developed for clinical decision support are typically validated using aggregate performance metrics such as the area under the receiver operating characteristic curve (AUC), sensitivity, and specificity, computed over an entire evaluation cohort. These global metrics can mask substantial heterogeneity in model reliability across individual patients and across subgroups of patients \citep{10.1145/3368555.3384468, 10.1145/3219819.3219930}. 
\textit{MED3pa} \citep{lefebvre_predictive_2026} addresses this by estimating, for a fixed and already-trained classifier (\textit{BaseModel}), a confidence score for each individual patient and for profiles of patients, which are groups of patients defined by shared characteristics and with similar confidence levels. This identifies \textit{at-risk-of-error profiles} with an interpretable tree structure, where the BaseModel exhibits reduced reliability and its predictions should be interpreted with caution.
This approach, however, does not indicate why a given profile is affected. Potential causes include irreducible aleatoric variability, epistemic uncertainty due to underrepresentation during training, or a distribution shift between the training and deployment populations \citep{kendall_what_2017, lahoti_responsible_2023}.


Another line of work targets the detection of distributional mismatch between the BaseModel's training and deployment contexts, helping ensure the model operates only within a validated deployment distribution \citep{silva_strategies_2025, quinonero-candela_dataset_2008}.
The \textit{Detectron} \citep{ginsberg_learning_2023} focuses on detecting \textit{harmful covariate shifts} (HCS), defined as any change in the covariate distribution between the training and deployment contexts that may weaken the generalization of a predictive model. Not every change in the input distribution degrades performance, as models may generalize beyond the exact support of their training distribution.
The Detectron therefore ties HCS detection directly to the BaseModel's behavior, detecting only those shifts in the covariate distribution that affect its generalizability.
Furthermore, despite that it is sample-efficient — requiring as few as ten test examples — the Detectron is designed and validated as a whole-population test and cannot determine whether a detected shift is confined only to a specific subgroup of patients.


These two methods are complementary as MED3pa identifies where a model underperforms and groups patients with similar predictive confidence using interpretable rules, while the Detectron can test if these localized failures are associated with a harmful distributional shift.
A closely related work is \textit{SHIFT} \citep{singh_who_2025}, which detects heterogeneous performance drift between training and deployment contexts and infers subgroup structure from the discovered shift patterns, which can lead to discrepancies between the subgroups flagged as shifted and the interpretable rule-defined subgroups \citep{lakkaraju_interpretable_2016, singh_who_2025}.
In contrast, our framework characterizes heterogeneous model reliability across patients, and performs harmful covariate shift detection within the actual rule-defined subgroups, thereby ensuring that the diagnostic test is applied precisely within interpretable regions of the population.

In this work, we therefore propose to use MED3pa's interpretable reliability profiles as localized regions in which Detectron can test for harmful covariate shift. This reveals shifts that are undetectable at the population level while distinguishing localized from population-wide degradation.
We evaluate this combined framework in a controlled setting where the ground-truth shift is known by design, across three clinical prediction tasks and three injected subgroup shifts.

\section{METHODS}

\paragraph{Data and tasks.}
We used data extracted from the MIMIC-IV \citep{johnson_mimic-iv_2023} database, combining chart and laboratory events, and patient demographics. From this dataset, three prediction tasks were defined:
\textbf{(i) 30-day readmission}, whether a patient is readmitted to the hospital within 30 days of a prior discharge; 
\textbf{(ii) 48-Hour live discharge}, whether a patient admitted to the intensive care unit (ICU) will be discharged alive within the following 48 hours; and \textbf{(iii) 48-Hour Mortality}, whether a patient admitted to the ICU will die within the following 48 hours. 
Each task-specific cohort was split in half into separate training and testing sets for the BaseModel using a patient-level split, ensuring that no patient's data contributed to both sets.
The training set was used to develop the BaseModel, and the testing set was used to simulate the detection of covariate shift.
For each task and shift condition, a separate BaseModel was trained after removing the corresponding subgroup from the training set, while the testing set remained unchanged.
Each BaseModel is a multi-layer perceptron (MLP) trained on its task-specific cohort, the pipeline is however model-agnostic and compatible with any classifier.





\paragraph{Injected covariate shift.} 
For each task, we construct three shifted training cohorts, each obtained by completely removing a single subgroup from training only: \textbf{(i) patients under 50} (age$<$50); \textbf{(ii) female patients} (sex$\neq$male); and \textbf{(iii) non-White patients} (race$\neq$white). The testing set is identical and representative across conditions, so any performance change is attributable to the induced training-time shift.

\paragraph{Profile discovery (MED3pa).}
For each combination of task and shift-condition, we first split the testing set in half. 
MED3pa is applied to the BaseModel using the first half, where it learns to estimate a per-patient confidence score that reflects the reliability of the BaseModel for each sample.
Using these confidence estimates on that same subset, a regression tree then partitions patients into interpretable, rule-based profiles grouping those with similar confidence scores.
The other half of the testing set is finally used to evaluate the BaseModel's performance within each profile.
MED3pa therefore identifies patient profiles defined by interpretable, feature-based rules in which the BaseModel exhibits systematically higher or lower performance.

\paragraph{Profile-level HCS detection (Detectron).}
For each patient profile $S$ identified by MED3pa, we use Detectron to determine whether it is affected by a harmful covariate shift between the BaseModel's training distribution $P$ and the testing distribution $Q$, so that a detected shift can be interpreted as a potential explanation for the performance gap observed in that profile.

Detectron identifies harmful shifts by learning an ensemble of \textit{constrained disagreement classifiers} (CDCs), which are initialized from the BaseModel and trained to preserve the same predictions on most of the training data ($P^t$), while also maximizing their disagreement with the BaseModel on both the testing distribution and a small reference subset ($P^r$) drawn from the training distribution ($P^t \cup P^r = P$, $P^t \cap P^r = \emptyset$). 
The three subsets $P^t$, $P^r$, and $Q$ allow Detectron to compare the disagreement that CDCs can induce on a controlled reference subset of the training distribution against disagreement on the testing distribution. 
Here, disagreement refers to the fraction of samples on which a CDC’s prediction differs from that of the BaseModel.  
If the testing distribution reflects a shift that alters the model's generalization behavior, the CDCs can more easily disagree with the BaseModel on the testing samples than on the reference subset. 
Detectron then determines whether this difference in disagreement is statistically significant, thereby detecting harmful covariate shift between the training and testing distributions.

\begin{figure}[ht]
    \centering
    \includegraphics[width=0.8\linewidth]{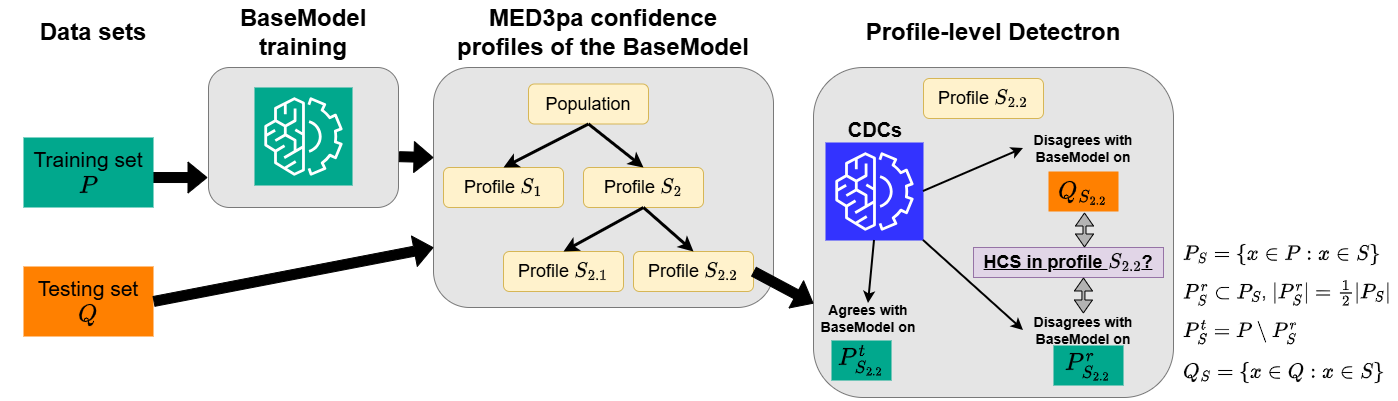}
    \caption{Overview of the profile‑level Detectron workflow. The BaseModel is trained on the full training distribution $P$ and evaluated on the testing distribution $Q$. MED3pa partitions the population into confidence‑based profiles. For each profile $S$, Detectron is applied by training constrained disagreement classifiers (CDCs) to match the BaseModel on the profiled training subset $P_S^t$ while inducing disagreement on the profiled testing subset $Q_S$ and reference subset $P_S^r$. A statistically significant difference in disagreement indicates harmful covariate shift (HCS) within that profile.}
    \label{fig:overall1}
\end{figure}

For each profile $S$, we extract $P_S = \{x \in P : x \in S\}$ and $Q_S = \{x \in Q : x \in S\}$. $P_S$ is split in half: one half, $P_S^r$, is held out as a profiled reference set, while the other half is combined with all out-of-profile training samples to form the profiled-training set $P_S^t = P \setminus P_S^r$, giving the CDCs enough data to preserve the BaseModel's behavior while retaining an in-profile reference for evaluating disagreement.
The CDCs' disagreement on $Q_S$ is then compared against their disagreement on $P_S^r$: if $Q_S$ reflects a shift that alters the BaseModel's generalization behavior, the CDCs can more easily disagree with it on $Q_S$ than on $P_S^r$. A statistically significant difference indicates HCS within the profile, which is assessed using a one-sided test with a false‑positive rate of $\alpha=0.05$ for declaring harmful covariate shift. We restrict this analysis to profiles containing at least 5\% of the overall population to ensure reliable statistical testing and to limit the number of profile-level tests, as runtime scales directly with the number of profiles. The full pipeline was run on approximately 10,000 samples using roughly 140 CPU cores over one week. 
Figure~\ref{fig:overall1} summarizes the full pipeline.
Code to reproduce all experiments is available at \url{https://github.com/MEDomicsLab/study_MED3pa_SHIFT2026}.

\section{RESULTS}

\textbf{Age shift.}
The injected age shift was not detected globally for any of the three tasks. Profile-level Detectron nevertheless identified HCSs within specific profiles, showing that a shift too weak to be detected at the population level can still be revealed within a localized region of the population.
For \textit{48-hour live discharge}, no HCS was detected in profiles covering 39.2\% of the population. The strongest localized effect occurred in patients with a Glasgow Coma Scale (GCS) Verbal  $\leq 3$, for whom sensitivity decreased to zero. 
A separate profile (GCS-Verbal $>$ 3, white blood cell count $>$ 14.88$\times 10^9$/L, hemoglobin $>$ 10.3g/dL) also exhibited an HCS. In this profile, the outcome prevalence increased from 9.47\% in training to 16.35\% in testing, a plausible explanation of the detected HCS despite the absence of performance degradation. By contrast, a closely related profile with lower hemoglobin ($\leq$ 10.3g/dL) showed no HCS, no change in prevalence, and no degradation in performance.
For \textit{48-hour mortality}, only one profile covering 8.76\% of the population was identified as affected by an age-related HCS: patients with GCS-Motor $\leq 5$ and age $\leq 66$. This profile showed a substantial decrease in sensitivity and specificity. No HCS was detected in the remaining profiles, which generally exhibited improved specificity in testing.
A similar localized effect was found for \textit{30-day readmission}. Although no population-level HCS was detected, Detectron flagged a profile covering 17.68\% of the population, characterized by hemoglobin $\leq 10.91$g/dL, length of stay $\leq 5$ days, and no specified malignant neoplasm. The mean age in this profile decreased from 73.3 to 67.8 years in between training and testing, consistent with the direction of the injected age shift. This profile also exhibited decreases in AUC (0.80 to 0.64), balanced accuracy (0.72 to 0.58), and sensitivity (0.67 to 0.42), while specificity remained relatively stable (0.77 to 0.75). Thus, although the age shift was not harmful at the population level, it was associated with a substantial performance degradation within this specific low-confidence profile.

\textbf{Sex shift. }
The sex-based injected shift was detected across all profiles for both the \textit{48-hour live discharge} and \textit{48-hour mortality} tasks. Performance degradation between training and testing was consistent across all profiles, with AUC and sensitivity decreasing by approximately 0.05-0.07, while specificity remained stable, indicating that this shift primarily affected the model's ability to identify positive outcomes.
For \textit{30-day readmission}, an HCS was detected in all but two profiles (23.8\% of the population), both corresponding to patients with prolonged stays ($>$7 days), but one with low hemoglobin ($\leq$11.6g/dL) and hematocrit ($\leq$30\%) while the other with higher hemoglobin ($>$11.6g/dL).
Despite not triggering Detectron, these profiles exhibited the same severe performance collapse as the rest of the population: AUC and sensitivity dropped by roughly 0.25, and specificity by roughly 0.10. This indicates that the BaseModel performs uniformly poorly across all patient profiles under this shift, with no subgroup showing resilience.


\textbf{Race shift.}
Race-related HCS was detected in every profile across all three tasks, accompanied by consistent decreases in AUC and sensitivity across profiles, while specificity remained comparatively stable.  This shows that this shift impacts the population similarly to the sex-related shift. 

\vspace{-0.08em}

\section{DISCUSSION \& CONCLUSIONS}

\vspace{-0.08em}

Our results demonstrate the complementarity of MED3pa and Detectron for diagnosing heterogeneous model reliability. Applying Detectron within MED3pa profiles revealed harmful shifts not apparent at the population level. 
In particular, the age shift (patients under 50, $\sim$18\% of the cohort) was undetected globally yet identified within specific profiles showing substantial performance degradation. In contrast, sex- and race-related shifts affect larger subgroups ($\sim$46\% female, $\sim$33\% non-White), producing broader effects across most or all profiles that were less informative for localizing the most affected patients.

Importantly, a detected HCS does not by itself explain the observed performance degradation. 
Some profiles showed performance decreases without a corresponding HCS (e.g., the two hemoglobin/hematocrit profiles under sex shift for 30-day readmission), while other detected HCS that did not necessarily translate into performance differences (GCS-Verbal $>$ 3, white blood cell count $>$ 14.88$\times 10^9$/L, hemoglobin $>$ 10.3g/dL under age shift for 48-hour live discharge). Furthermore, because Detectron is applied independently within each profile at $\alpha$ = 0.05, no correction for multiple comparisons is applied, and results should be interpreted as exploratory signals rather than confirmatory findings. 
Profiles showing a detected HCS and performance degradation represent particularly strong candidates for further investigation or targeted model adaptation.

This study is a proof‑of‑concept with several limitations. The experiments rely on controlled single-axis subgroup shifts injected into a single dataset, which simplifies the evaluation but does not capture the diversity or complexity of naturally occurring deployment shifts. In addition, we only evaluate shifts that affect relatively large portions of the population; future work should examine whether profile‑level HCS detection remains effective when shifts impact much smaller subgroups. Broader validation on additional clinical datasets and under more realistic temporal, institutional, and demographic shifts is also needed. 
Several pipeline design choices, including the reference-set ratio and the minimum profile size, were fixed a priori and could be optimized in future work. The pipeline's diagnostic accuracy, its profile-level false-positive rate, and how it compares with subgroup-monitoring methods like \textit{SHIFT} \citep{singh_who_2025} remain to be evaluated.

Overall, combining MED3pa with profile-level Detectron extends reliability analysis from identifying \textit{where} a model is less reliable to investigating whether a harmful distribution shift may explain \textit{why} it is less reliable in those patient profiles.

\bibliography{iclr2025_conference}
\bibliographystyle{iclr2025_conference}


\end{document}